\documentclass[letterpaper, 10 pt, conference]{ieeeconf}  

\IEEEoverridecommandlockouts                              

\usepackage{graphicx} 
\usepackage{mathptmx} 
\usepackage{times} 
\usepackage{amsmath} 
\usepackage{amssymb}  
\let\mathcal\undefined
\DeclareMathAlphabet{\mathcal}{OMS}{cmsy}{m}{n}  

\usepackage{array}
\usepackage{verbatim}

\usepackage{flushend}  

\usepackage{hyperref}
\usepackage{multirow}
\usepackage{xfrac}

\usepackage{subcaption}

\usepackage{xcolor} 
\usepackage{color,soul}
\usepackage{ulem}
\usepackage{adjustbox}

\newif\ifversionwithcorrections      \versionwithcorrectionstrue %
\ifversionwithcorrections

\newcommand{\removedtext}[1]{\adjustbox{bgcolor=red!30,varwidth={\linewidth}}{\st{#1}}}
\newcommand{\rremovedtext}[1]{\adjustbox{bgcolor=red!30,varwidth={\linewidth}}{\hsout{#1}}}

\else

\newcommand{\removedtext}[1]{}
\newcommand{\rremovedtext}[1]{}

\fi

\title{\LARGE \bf
Road Scene Graph: A Semantic Graph-Based Scene Representation Dataset for Intelligent Vehicles 
}
\author{Yafu Tian$^{1,4}$, Alexander Carballo$^{2,3}$, Ruifeng Li$^{4}$ and Kazuya Takeda$^{1,2,3}$
\thanks{$^{1}$ Graduate School of Informatics, Nagoya University, Furo-cho, Chikusa-ku, Nagoya 464-8603, Japan.}%
\thanks{$^{2}$ Institute of Innovation for Future Society, Nagoya University, Furo-cho, Chikusa-ku, Nagoya 464-8601, Japan.}%
\thanks{$^{3}$ Tier IV Inc., Nagoya University Open Innovation Center, 1-3, Mei-eki 1-chome, Nakamura-Ward, Nagoya, 450-6610, Japan.}%
\thanks{$^{4}$ State Key Laboratory of Robotic and Intelligent System, Harbin Institute of Technology, Heilongjiang, China. \hfill\break%
Email:{\tt\scriptsize\underline{yafu.tian@g.sp.m.is.nagoya-u.ac.jp}} (corresponding author), \tt\scriptsize{alexander@g.sp.m.is.nagoya-u.ac.jp, lrf100@hit.edu.cn, 
kazuya.takeda@nagoya-u.jp.}}}%

\begin{document}

\maketitle
\thispagestyle{empty}
\pagestyle{empty}

\begin{abstract}

Rich semantic information extraction plays a vital role on next-generation intelligent vehicles.
Currently there is great amount of research focusing on fundamental applications such as 6D pose detection,
 road scene semantic segmentation, etc. And this provides us a great opportunity to think about how shall these data be organized and exploited. 

In this paper we propose road scene graph,
 a special scene-graph for intelligent vehicles. Different to classical data representation, 
 this graph provides not only object proposals but also their pair-wise relationships. 
 By organizing them in a topological graph, these data are explainable, fully-connected, and could be easily processed by GCNs (Graph Convolutional Networks). Here we apply scene graph on roads using our Road Scene Graph dataset\footnote[5]{The road scene graph dataset is based on nuScenes\cite{caesar2020nuscenes} dataset and CARLA simulator \cite{Dosovitskiy17}.}, including the basic graph prediction model. This work also includes experimental evaluations using the proposed model.
 \addtocounter{footnote}{5}  
\end{abstract}



\section{Introduction}
\label{s:introduction}
\begin{figure}[htbp]
        \centering
        \includegraphics[width=1\linewidth]{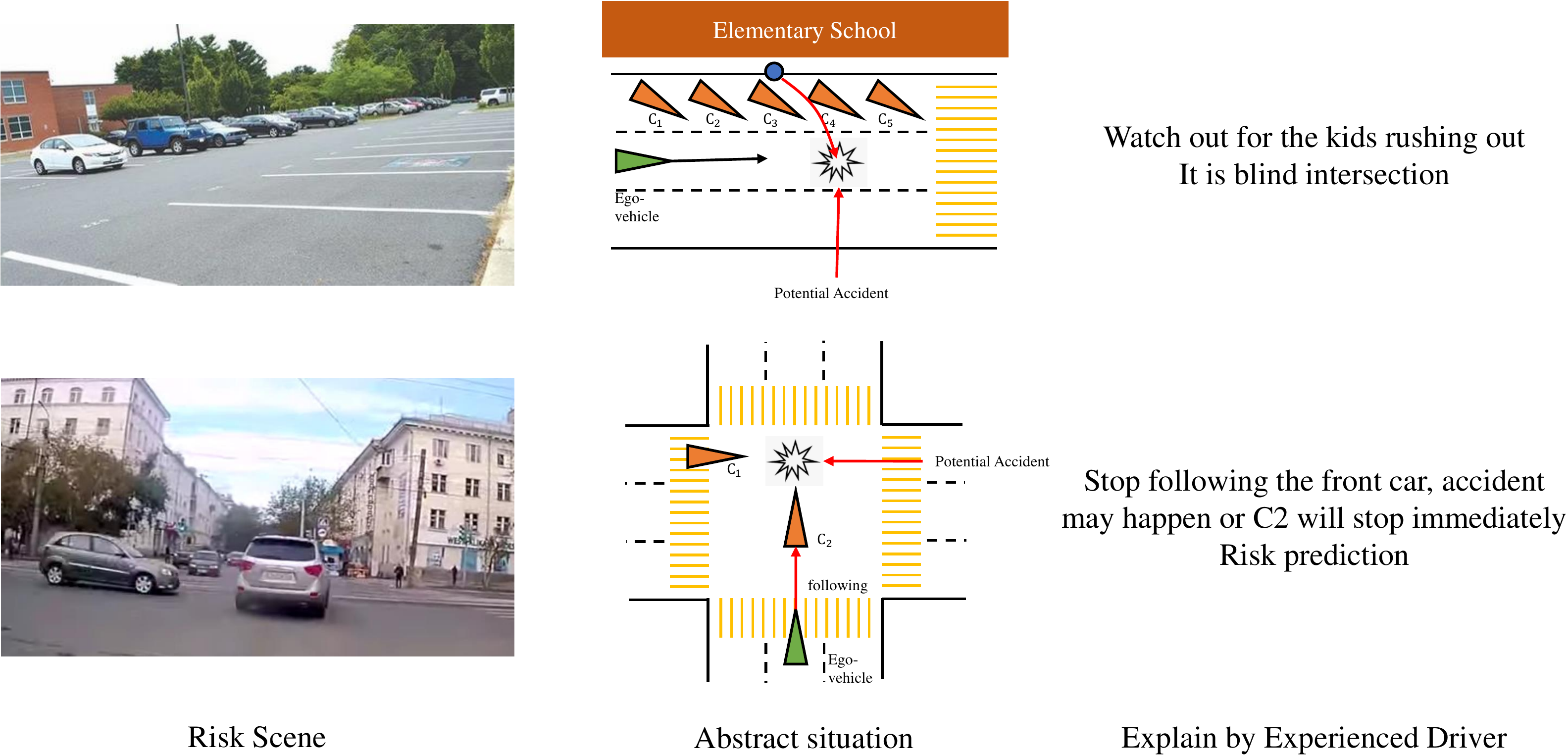}  
        \caption{Samples of hidden risks, hard to predict by self-driving vehicles, but simple for humans.}
        \label{introExample}
\vspace{-1em}
\end{figure}

The area of self-driving vehicles has been deeply researched and widely developed. However, there is still a great performance gap between human driver and AI systems, this because an experienced human driver can fully utilize rich semantic information from the ego-vehicle. To predict a series of potential risks, like blind intersections, kids rushing out, other vehicles, or even inexperienced drivers on the highway, is still a hard task for automated vehicles. Most of the potential risks are hidden in the semantic and behavior level, and based on driver's experience/common sense. Fig.~\ref{introExample} illustrates a couple of examples for these kinds of hidden risks, and an explanation given by a human driver. 

So far, the application of road semantic information is too simple to predict such hidden risks. Therefore, new scene representations, other than bounding boxes and attention masks, becomes necessary for intelligent vehicles.

\begin{figure}[htbp]
        \centering
        \includegraphics[width=1\linewidth]{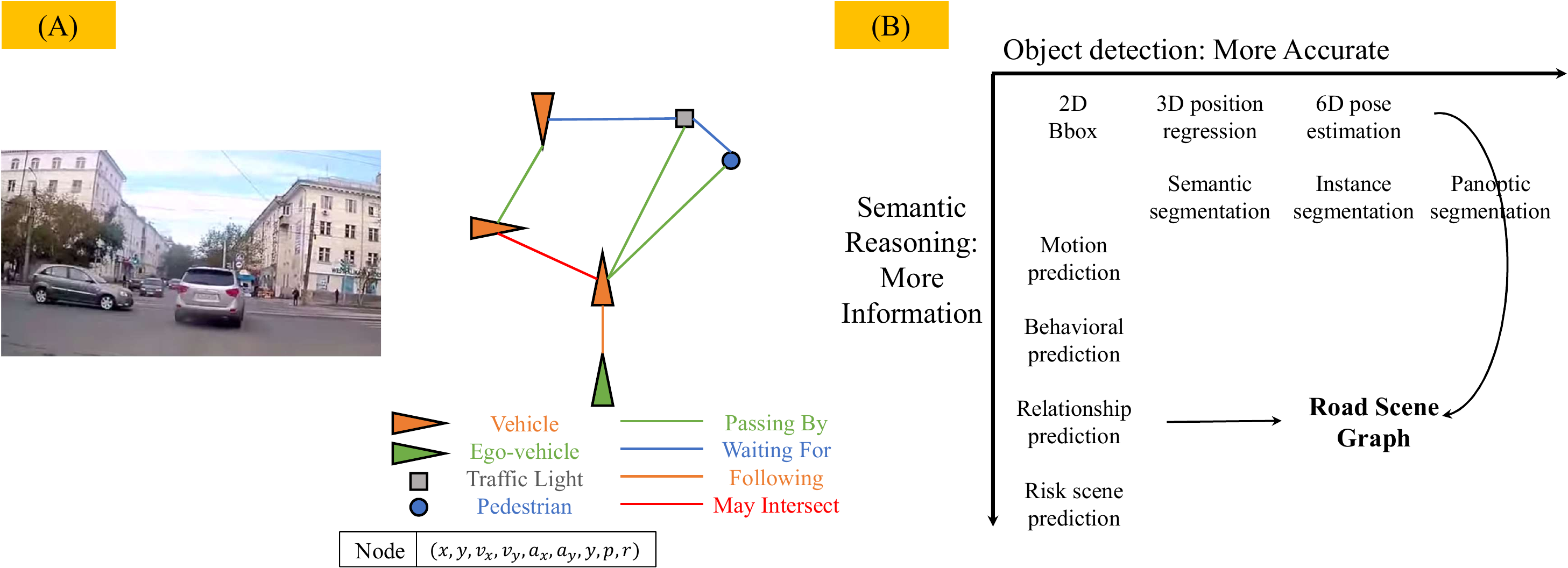}  
        \caption{(A) example of Road Scene Graph, (B) relationship with other environment recognition methods.}
        \label{introExample1}
\vspace{-0.5em}
\end{figure}

Scene graph, as a widely-used structured data representation and computation method, quite powerful in common image understanding area, proposes a better alternative to deal with the complexities of the real world, and it is located between model-based and end-to-end deep network model. As Fig.~\ref{introExample1}(a) illustrates, in scene graph objects are encoded as nodes; their relationships, such as "following" and "waiting", represented as edges. And Fig.~\ref{introExample1}(b) shows the relationship between road scene graph and other environment recognition methods. Road scene graph combines bounding box regression and behavior/relationship prediction, so it also benefits from the rapid development of object detection methods and behavior prediction models. 

From the scene structural aspect, traditional methods, such as bounding boxes and segmentation masks, only focus on objects and pay little attention to their relationships, which could be of vital importance for vehicle's safety. And from the method and algorithm's point of view, currently many recognition and perception modules are implemented using end-to-end convolutional networks. So the "detection" and "inference" parts are actually hidden in a black box. Road scene graph is a explainable "bridge" between these two modules.

Also, scene graph benefits from the current progress on Graph Convolutional Networks (GCN), and achieves significant advantages in many tasks such as: scene  understanding, visual-language navigation, visual-question answering, object  detection, and so on. However, most of these areas are about common objects and their relationships. Our research work applies scene graph methods to intelligent vehicles. We propose here the Road Scene Graph dataset which includes more than 500 real-world driving scenes\footnote{These scenes, and geometry informations come from nuScenes dataset\cite{caesar2020nuscenes} and CARLA Simulator \cite{Dosovitskiy17}.}, and their corresponding road scene graphs. Firstly, we use a model-based system to manually annotate and generate a simple road scene graph dataset. Then, this dataset is used to train a GCN to predict scene graph for unknown scenes. This prediction has been used as an assistant for manual data annotating process.

The contributions of this work are summarized as follows:

\begin{itemize}
        \item We apply the concept of scene graph to the area of intelligent vehicles, and propose a set of criteria for generating road scene graphs.
        \item We propose Road Scene Graph dataset, which is an intelligent-vehicle-oriented scene graph dataset with more than 500 annotated scenes.
        \item We propose a set of models to predict missing edges in road scene graph, and then generate such scene graph from object proposal.
\end{itemize}

This paper is structured as follows: Section~\ref{s:relworks} presents related datasets featuring LiDARs, while Section~\ref{s:roadscene} describes our Road Scene Graph Dataset. Then in Section~\ref{s:roadgraphnetwork}, we propose simple relationship refinement and prediction model as a baseline. Experimental results in Section~\ref{s:experiment}, showing that the road scene graph could be a perfect structure to organize rich semantic data in driving scenes. Finally, this paper is concluded in Section~\ref{s:conclusions}.


\section{RELATED WORK}
\label{s:relworks}

\subsection{Capturing Semantic Relationship Data for Intelligent Vehicles}
As Fig.~\ref{introExample1} (B) illustrates, most environment recognition research for intelligent vehicles were guided by two doctrines. Firstly, many research focusing on more accurate geometry information around ego-vehicle. It starts from simple object detection tasks in early 2000s, to very accurate object detection and 6D Pose estimation \cite{kuo2015deepbox, cho2014multi, xu2018pointfusion}, semantic, instance and panoptic segmentation masks. And it proposes real-time safety for all detected objects. 

Another direction is about semantic reasoning, which focuses on the safety of a little bit longer time, about 5-20 seconds, and the potential risk which could not be percepted by geometry. It starts from simple trajectory prediction and vehicle behavioral analyzation\cite{li2019grip, altche2017lstm}.  And proves human-like behavior for an intelligent vehicle, and predict potential risk. Also, research \cite{kim2017interpretable, kim2018textual}  are focusing on the scene captioning, and intent explaining. So it brought out more interpretability to intelligent vehicles. Also, many kinds of research and datasets for pedestrian behavior prediction have been released \cite{rasouli2017they, kooij2019context} , bringing more safety insight for intelligent vehicles. 

So, an interesting idea is whether we can design a data structure, which could capture both accurate object detection and rich semantic information. And a powerful and effective method for processing these disparate data sources is to combine them into a graph \cite{wu2020comprehensive, battaglia2018relational} , and using graph neural network \cite{wu2020comprehensive, kipf2016semi, defferrard2016convolutional}  to extract potential patterns. 

\subsection{Scene Graph Generation and its Application}
Among existing research in intelligent vehicles, there are several works using hand-engineering model-based systems, and also a great amount of end-to-end driving models\cite{bojarski2016end, chen2017end, chen2020learning, carballo2018end}. Both methods archive amazingly good results on behavior and performance. However, when it comes to highly complex semantic information, combinatorial generalization shall be a top priority for an intelligent vehicle to achieve human-like response\cite{battaglia2018relational}. And graph-based scene representation is an excellent way to learn this generalization \cite{battaglia2018relational, zambaldi2018deep}. 

With the fast development of GCNs (Graph Convolutional Networks) \cite{kipf2016semi, defferrard2016convolutional} and graph generation models \cite{simonovsky2018graphvae, lu2016visual, xu2017scene} these years, several works are using scene graph to achieve state-of-art performance in a large variety of areas. This includes visual scene understanding tasks, visual-language navigation \cite{gupta2017cognitive, Yang2018VisualSN}, visual-question answering\cite{li2019relation, narasimhan2018out}, object detection, etc. There are also papers and datasets for scene-graph generation\cite{zellers2018neural, li2017scene, xu2017scene}. However, most of these are about common objects and their relationships\cite{krishna2017visual}. To the best of the authors' knowledge, we are the first group to apply scene-graph to traffic scenes and corresponding relationships.

\subsection{Rich Semantic Dataset for Intelligent Vehicles}
Data plays an important role in state-of-art data-driven research about intelligent vehicles. In the past few years, there are various datasets \cite{caesar2020nuscenes, wolfe2020rapid, behley2019semantickitti, sun2020scalability, yu2020bdd100k, kesten2019lyft, maddern20171, meyer2019automotive} and driving simulators\cite{Dosovitskiy17, airsim2017fsr}, with an unquestionable impact to research. In Table~\ref{relatedDatasets}, we present some of those datasets which include rich semantic information, such as images, bounding boxes, multiple labels, lanes, etc. Another important dimension is their degree of completeness, this is, whether the dataset considers multiple situations, including rain, snow, fog, daytime, nighttime, and so on. Also, risky scenes are important for intelligent vehicles safety research. 

The final goal of our research is to enable intelligent vehicles to recognize, predict, and even avoid potential risks. Currently there are few datasets focusing on this aspect. Although it is possible to simulate accidents using simulators like CARLA\cite{Dosovitskiy17} and Airsim\cite{airsim2017fsr}, risky scenes obtained from the real world are necessary. Works such as the Road Hazard Stimuli dataset\cite{wolfe2020rapid}, including about 500 accident/normal scenes taken from YouTube videos with annotations, are of the significant importance.

\begin{table*}[h]
	\centering
	\caption{Public road datasets with rich semantic information.}
	\includegraphics[width=0.7\linewidth]{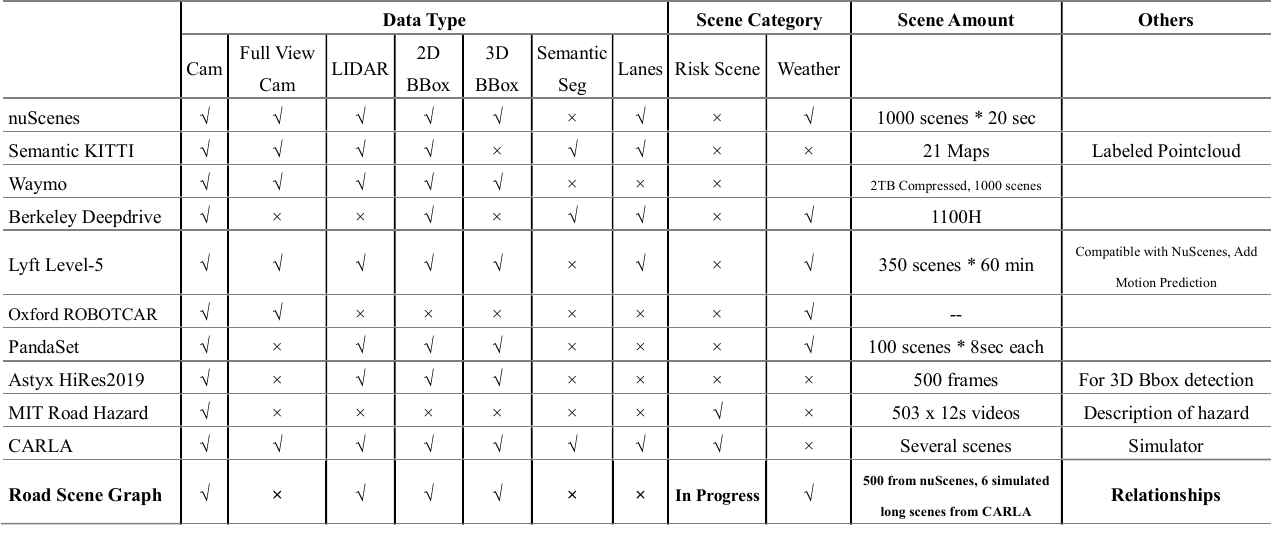}  
	\label{relatedDatasets}
	\vspace{-2em}
\end{table*}
\section{ ROAD SCENE GRAPH DATASET }
\label{s:roadscene}
In this work, we introduce our Road Status Graph dataset\footnote{Our Road Status Graph dataset will be made available at \url{https://github.com/tianyafu/road-status-graph-dataset}} 
 and how this dataset has been constructed. For common objects and their relationships, there are open datasets such as Visual Genome \cite{krishna2017visual} and VRR-VG \cite{liang2019vrr}. However, when it comes to intelligent vehicles area, the objects and relationships are different from common datasets. We built our own dataset based on previous datasets illustrated in Table~\ref{relatedDatasets}. The objects and their relationships we are modeling are listed in Table~\ref{relationships} and Table~\ref{objAndAttributes}.

Road scene graph is a multigraph $G = \langle{V,E}\rangle$, similar to a scene graph, includes two parts. The node set $V$, where each node $v_i \in V $ correspond to a feature vector representing objects such as vehicles, pedestrians, traffic lights, obstacle cones, etc. And the edge set $E$, where each edge, such as {\texttt{vehicle-waitingFor-pedestrian}}, $e_i \in E$ is a triplet $e_i = {(v_i, v_j, l_i) \; | \; v_i, v_j \in V, \; l_i \in L}$, where $L$ corresponds to the relationship types set. Edges in $E$ have two kinds of relationships. The typical relationship are established between two different objects, for example a "vehicle" passing by a "pedestrian". The attribute relationship models self-loop edges, it is used to describe an object's properties, and gives extra flexibility to our road scene graph. 

\begin{figure}[htbp]
        \centering
        \includegraphics[width=0.85\linewidth]{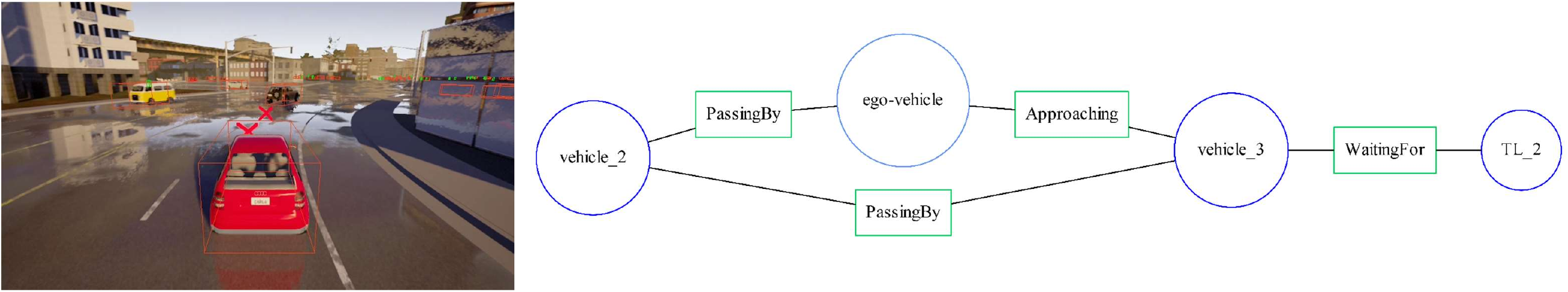}  
        \includegraphics[width=1\linewidth]{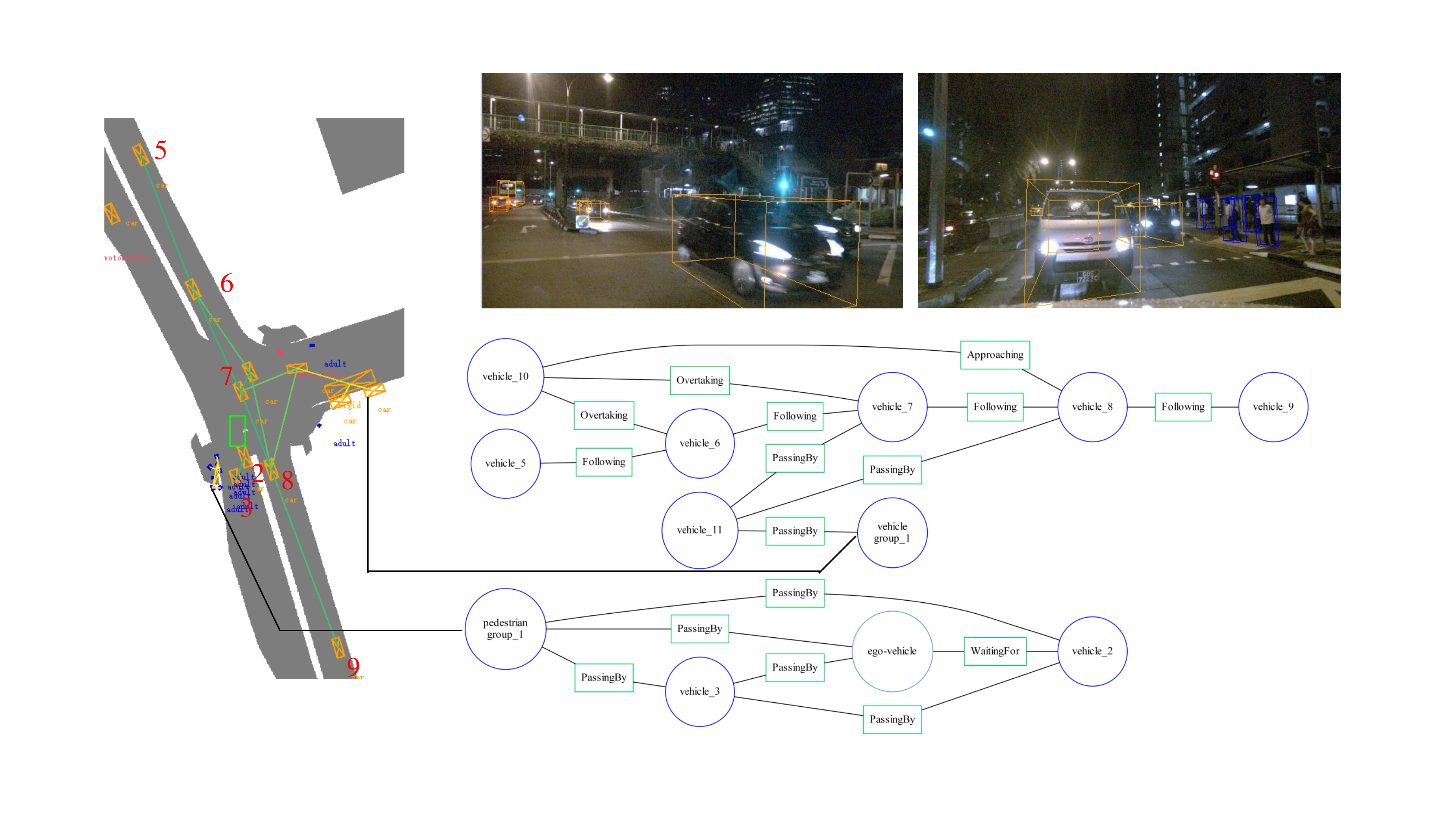}  
        \caption{Sample scenes in Road Scene Graph Datasets. Top row using CARLA \cite{Dosovitskiy17}, bottom row using nuScenes \cite{caesar2020nuscenes} and more complicated. Blue circles represent nodes, green squares in edges represent the relationship between objects.}
        \label{Datasetintro}
\vspace{-1em}
\end{figure}

Figure~\ref{Datasetintro} illustrates simple examples of our road scene graph dataset.\footnote{The bottom sample comes from nuScenes, scene-1100, 5.5s (sample 11) Please refer to this scene for its video context information.}. These samples comes from both CARLA simulator\cite{Dosovitskiy17} and nuScenes \cite{caesar2020nuscenes}, nuScenes provides a richer variety of driving scenes. CARLA allows to create configurable and flexible scenes, allowing researchers to generate condition-specific scenarios for better evaluation. 

Table~\ref{datasetProperty} lists some basic properties of the road scene graph dataset. nuScenes only provides short scenes (20 seconds each), thus we also publish some scenes from CARLA simulator, to give a long-time status graph sample.

\begin{table}[htbp]
        \caption{Data property for road scene graph dataset}
        \begin{tabular}{|c|c|c|}
        \hline
        Data Source         & Scene Amounts                                                        & Others                                                                          \\ \hline
        nuScenes            & \begin{tabular}[c]{@{}c@{}}500 scenes\\ 20 seconds each\end{tabular} & \begin{tabular}[c]{@{}c@{}}More complex scenes with\\ more objects \& relationships\end{tabular} \\ \hline
        CARLA               & 5 (for validation)                                                   & 4 of 5 mins and 1 of 20 mins                                                          \\ \hline
        \end{tabular}
        \centering
        \label{datasetProperty}
\vspace{-0.5em}
\end{table}

Table~\ref{relationships} lists the potential kinds of relationships \footnote{Relationships marked with a star '*' means they can be detected in CARLA with our implemented classifier, please refer to Section~V.} in road scene graph dataset. And Table~\ref{objAndAttributes} is for the objects and corresponding attributes in this dataset.

One type of relationship worth mentioning is the group. Pedestrians or vehicles nearby tend to move in the same mode thus forming a group or cluster. For example, when they waiting for a traffic light and passing through the crossroad, their speed, direction, and behavior stay relatively the same. In our Road Scene Graph Dataset, this is the only transitive relationship. By adding this kind of relationship, the annotation work becomes more efficient: when a vehicle is passing by a set of traffic cones, the annotator does not need to add multiple "Passing By" relationships to all cones, instead add a single "Passing By" relationship to the cone group.

\begin{table}[htbp]
        \caption{Objects and relationships in road scene graph dataset}
        \setlength{\tabcolsep}{5pt}
        \begin{tabular}{|c|c|c|c|c|}
        \hline
                                                                    & Human & Vehicle                                                                                                                        & Obstacle                                                      & Traffic-sign                                                                      \\ \hline
        Human                                                       & Group* & \begin{tabular}[c]{@{}c@{}}Behind*\\ on-lane*\\ waiting-for-cr\\ May-intersect\end{tabular}                                      & Behind*                                                        & waiting-ts*                                                                        \\ \hline
        \begin{tabular}[c]{@{}c@{}}Vehicle\\ (cyclist)\end{tabular} & --    & \begin{tabular}[c]{@{}c@{}}Group*\\ Same-lane*\\ Following*\\ Approaching*\\ Waiting-for-cr\\ Passing by*\\ Overtaking\end{tabular} & \begin{tabular}[c]{@{}c@{}}passing-by\\ avoiding\end{tabular} & \begin{tabular}[c]{@{}c@{}}waiting-for-ts*\\ stop-by-ts*\\ react-by-ts*\end{tabular} \\ \hline
        Obstacle                                                    & --    & --                                                                                                                             & Group                                                         & behind                                                                            \\ \hline
        Traffic-sign                                                & --    & --                                                                                                                             & --                                                            & --                                                                                \\ \hline
        \end{tabular}
        \centering
        \label{relationships}
\vspace{-1em}
\end{table}


\begin{table}[htbp]
	\caption{Objects and attributes in road scene graph dataset}
	\setlength{\tabcolsep}{4pt}
	\begin{tabular}{p{0.04cm}|c|c|c|c|}
		\cline{2-5}
		\multirow{1}{*}{\rotatebox{90}{\mbox{\scriptsize{Objects}}}} &
		\begin{tabular}[c]{@{}c@{}}Human\\ (Walker, kids\\ workman, etc.)\end{tabular} & 
		\begin{tabular}[c]{@{}c@{}}Vehicle\\ (cyclist)\end{tabular}  &
		\begin{tabular}[c]{@{}c@{}}Obstacle\\ (Barrier,\\ debris, etc.)\end{tabular} &
		traffic-sign \\ \cline{2-5}
		
		\multirow{1}{*}{\rotatebox{90}{\mbox{\scriptsize{Attributes}}}} &
		\begin{tabular}[c]{@{}c@{}}Stop\\ Moving\\ near-crossroad\\ near-lane\end{tabular} &     
		\begin{tabular}[c]{@{}c@{}}Go-stright\\ Accleration\\ Slow-down\\ Parking\\ turn-left\\ turn-right\\ intersection-passing\\ left/right-lane-branch\\ left-right-lane-changing\\ passing-cr\\ u\_turn\end{tabular} &
	    \begin{tabular}[c]{@{}c@{}}On-lane\\ On-roadside\end{tabular} &
	    --\\ \cline{2-5}
	\end{tabular}
	\centering
	\label{objAndAttributes}
\vspace{-1em}
\end{table}


\section{ ROAD SCENE GRAPH PREDICTION NETWORK }
\label{s:roadgraphnetwork}
In this section, we propose 3 basic graph generation models for baseline evaluation. Each model is designed for a specific task:

\begin{enumerate}
        \item  Road scene graph refinement network (RSGRN): randomly delete $k$ edges in the graph, and the task is to predict those deleted edges.
        \item  Next graph prediction using graph VGAE (NGPGV): given road scene graph in time $t$, predict the graph for $t+1$.
        \item  Next graph prediction using graph VGAE and bird's-eye view (NGPGVEB): similar to task (2), but adding CNN encoded bird's-eye view to represent obstacles, roads, and lanes.
\end{enumerate}

\begin{figure}[htbp]
        \centering
        \includegraphics[width=1\linewidth]{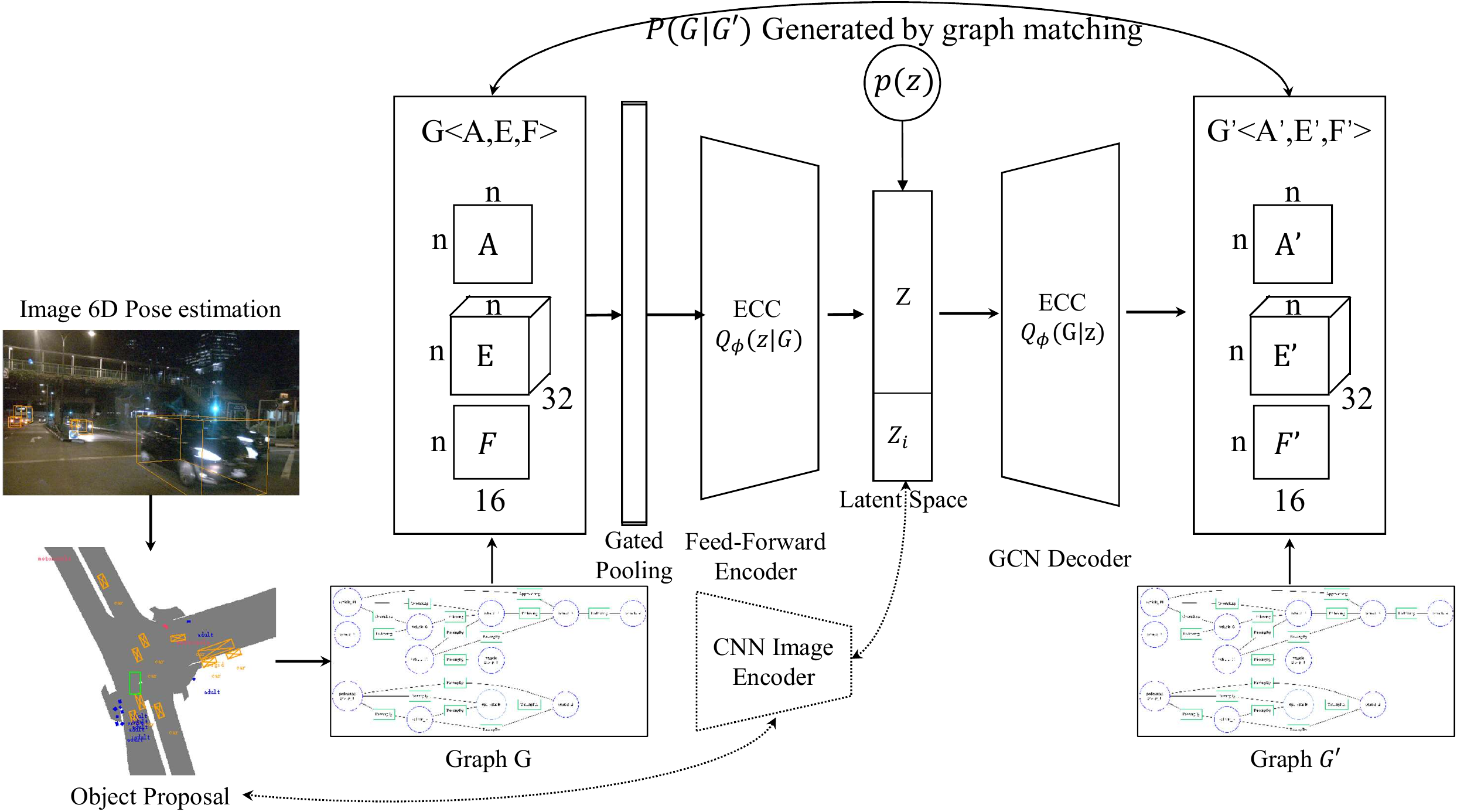}  
        \caption{Our initial model to predict pairwise relationships between objects, and find missing relationships in the graph.}
        \label{model}
\vspace{-1em}
\end{figure}

For these tasks, our model, shown in Fig.~\ref{model}, uses variational graph auto encoder (VGAE) and is based on the works from Simonovsky~et al.\cite{simonovsky2018graphvae} and Kipf~et al.\cite{kipf2016variational}. The model proposed in \cite{simonovsky2018graphvae} is effective for generating small graphs (Graph size $k \leqslant 39$). 

In our model, the graph $G = \langle{A,E,F}\rangle$ is represented by three components: adjacency matrix $A$, edge embedding tensor $E$ and node embedding vector $F$. The nodes in $F$ are manually defined: $v_i = \langle{\mathbf{l}_{ni}, x_i, y_i, v_{xi}, v_{yi}, a_{xi}, a_{yi}, y, p, r}\rangle$, where $\mathbf{l_{ni}}$ represents the one-hot label of node's type, position ($x_i$, $y_i$),  velocity ($v_{xi}$, $v_{yi}$), acceleration ($a_{xi}$, $a_{yi}$), and heading angles (yaw $y$, pitch $p$, and roll $r$). Our model includes four submodels: Gated pooling, Feed-forward Encoder, GCN Decoder and fast graph matching algorithm.

The feed-forward pooling is based on \cite{simonovsky2017dynamic}. The edge-conditioned convolution is formalized as follows:

\begin{equation}
        X^l(i) = \frac{1}{|N(i)|} \sum_{j \in N(i)} F^l(L(j,i);w^l)X^{l-1}(j) + b^l
\end{equation}

We used a global gated pooling, based on \cite{li2015gated}, instead of adding pooling between each layer because the size of the road scene graph is limited. 

Based on the original graphVAE model in \cite{simonovsky2018graphvae}, the reconstruction loss is a weighted sum of adjacency loss, node feature loss and edge loss, as shown in Eq.~\ref{overallDecodingLoss}.

\begin{equation}
        \label{overallDecodingLoss}
\begin{aligned}
        \mathcal{L} = - \log p(G|z) =& -\lambda_A \log p(A'|z) - \lambda_F \log p(F|z) \\
         &- \lambda_E \log p(E|z)
\end{aligned}
\end{equation}

And the three losses were defined as follows. Let $A'$ be the 
adjacency matrix, $E'$ the edge embedding tensor and $F'$ the node embedding vector in the generated graph $G'$, shown in Fig.~\ref{model}. Also, let $\widehat{A'}=XAX^T$, $\widehat{F'}=X^TF'$, $\widehat{E'}_{.,.,l} = X^TE'_{.,.,l}X$:

\begin{equation}
        \begin{aligned}
                \log p(A'|z) =& \frac{1}{k} \sum_{a}\widehat{A}_{a,a} \log A'_{a,a} + (1-\widehat{A}_{a,a})\log (1-A'_{a,a}) \\
                &+\frac{1}{k(k-1)} \sum_{a \neq b} \widehat{A}_{a,b} + (1-\widehat{A}_{a,b}) \log (1-A'_{a,b})
        \end{aligned}
\end{equation}

\begin{equation}
        \log p(F|z) = \frac{1}{n} \sum_{i} \log F_i^T \widehat{F'}_{i,.} 
\end{equation}

\begin{equation}
        \log p(E|z) = \frac{1}{||A||_l - n} \sum_{i \neq j} \log E_{i,j,.}^T\widehat{E'}_{i,j,.}
\end{equation}

\section{IMPLEMENTATION AND EXPERIMENT}
\label{s:experiment}
\begin{figure*}[htbp]
        \centering
        \begin{subfigure}{.39\textwidth}
                \includegraphics[width=1\linewidth]{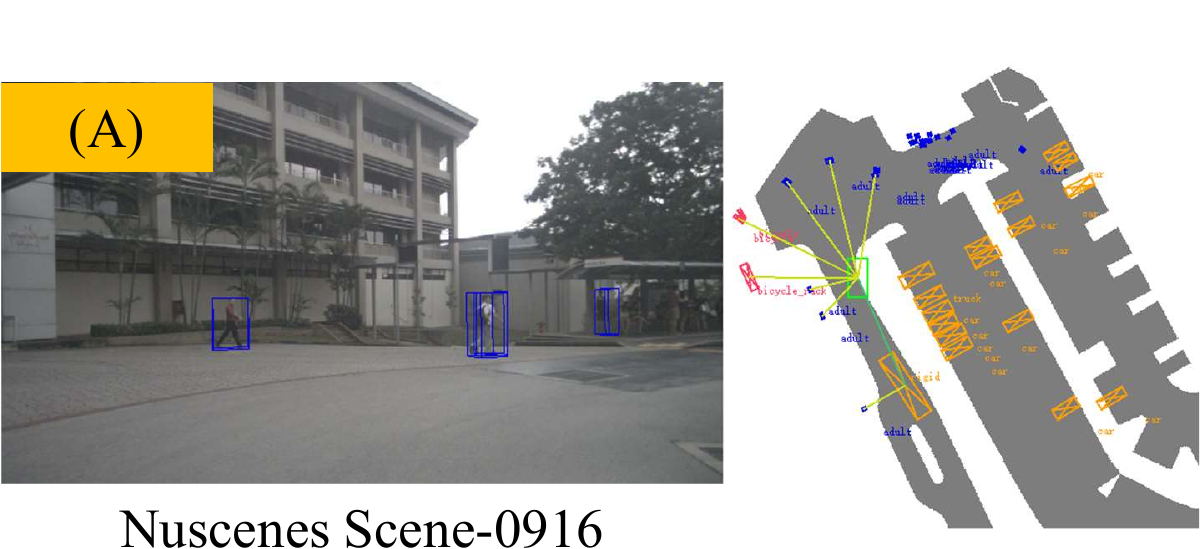}  
        \end{subfigure}
        \begin{subfigure}{.6\textwidth}
                \includegraphics[width=1\linewidth]{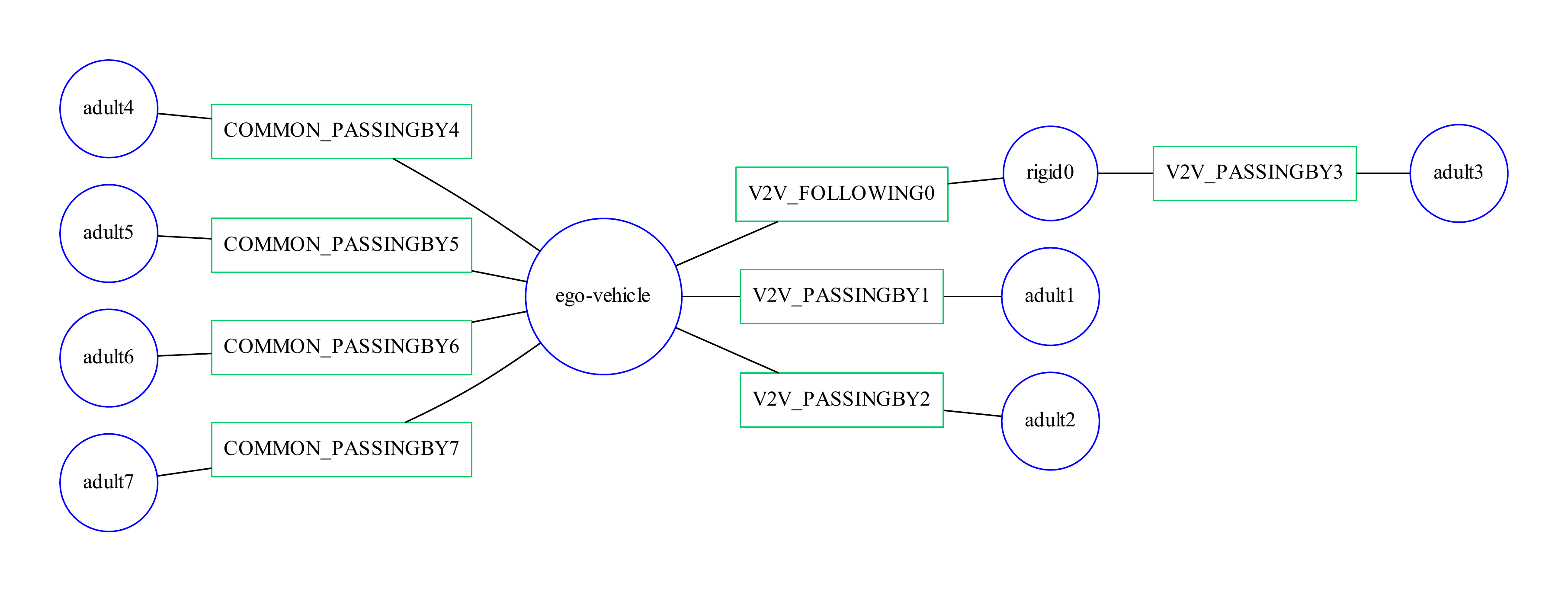}  
        \end{subfigure}
        \newline
        \begin{subfigure}{.39\textwidth}
                \includegraphics[width=1\linewidth]{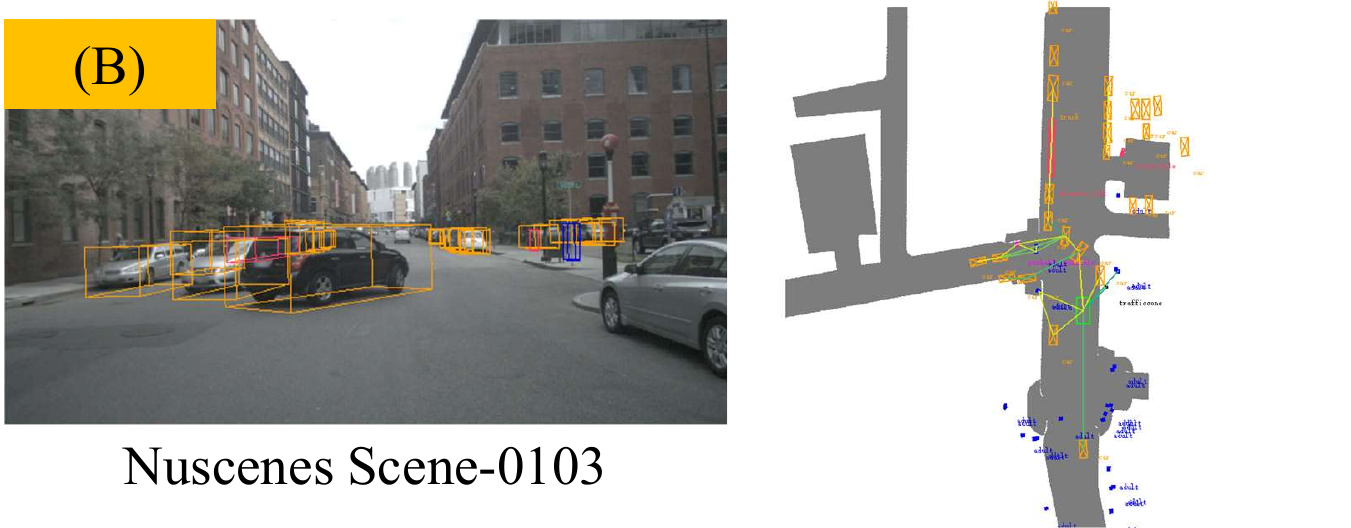}  
        \end{subfigure}
        \begin{subfigure}{.6\textwidth}
                \includegraphics[width=1\linewidth]{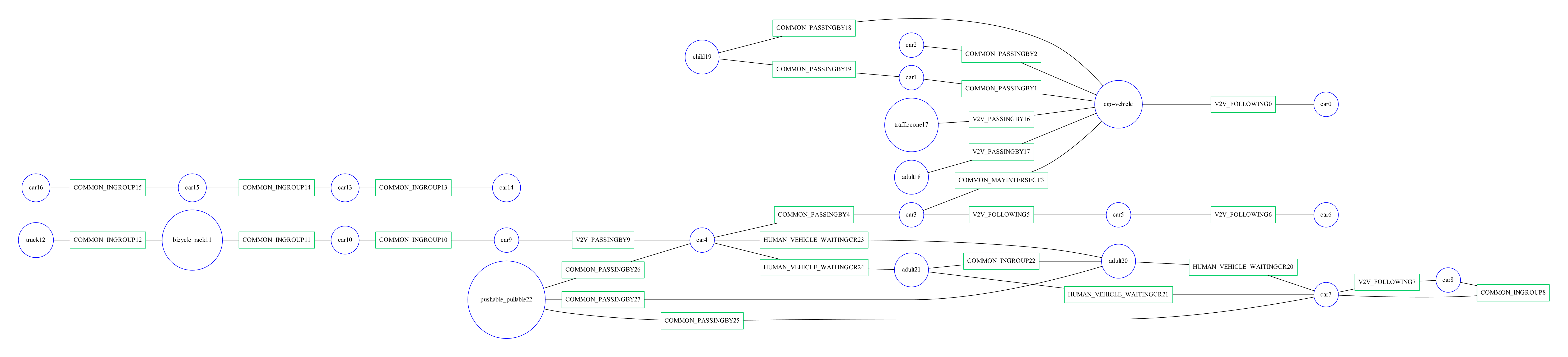}  
        \end{subfigure}
        \newline
        \begin{subfigure}{.39\textwidth}
                \includegraphics[width=1\linewidth]{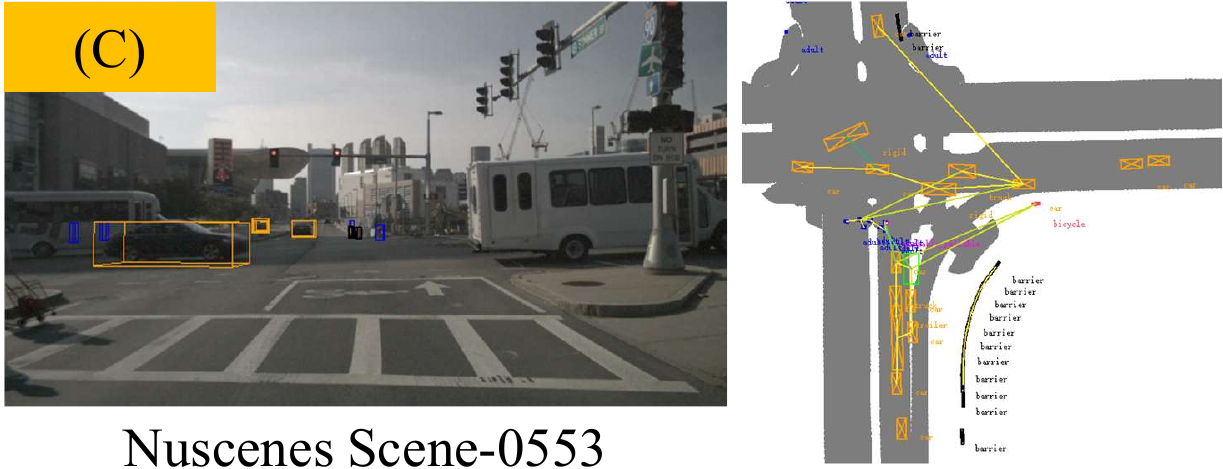}  
        \end{subfigure}
        \begin{subfigure}{.6\textwidth}
                \includegraphics[width=1\linewidth]{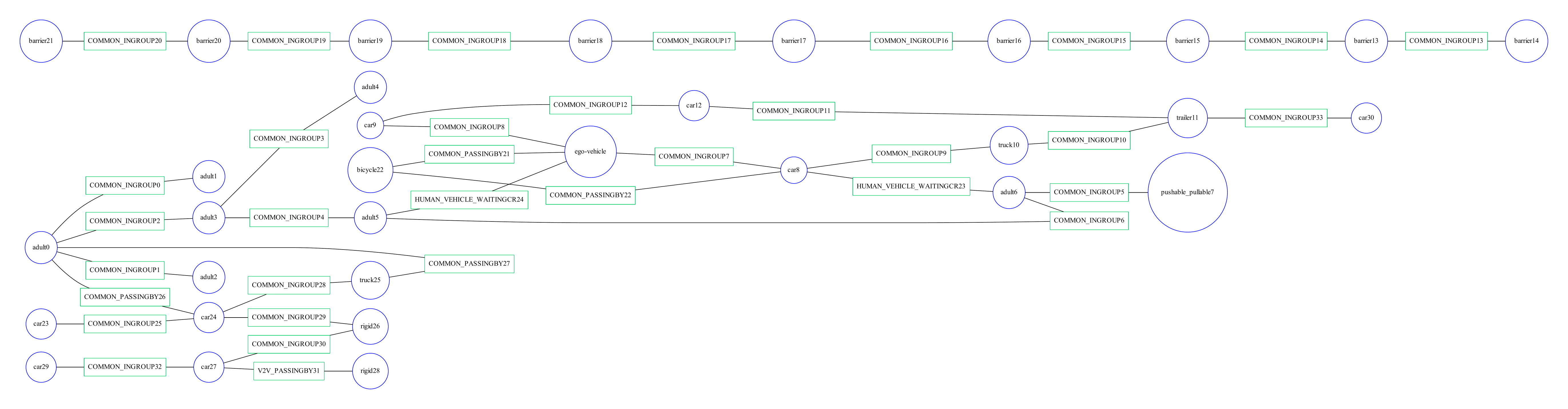}  
        \end{subfigure}
        \newline
        \begin{subfigure}{.39\textwidth}
                \includegraphics[width=1\linewidth]{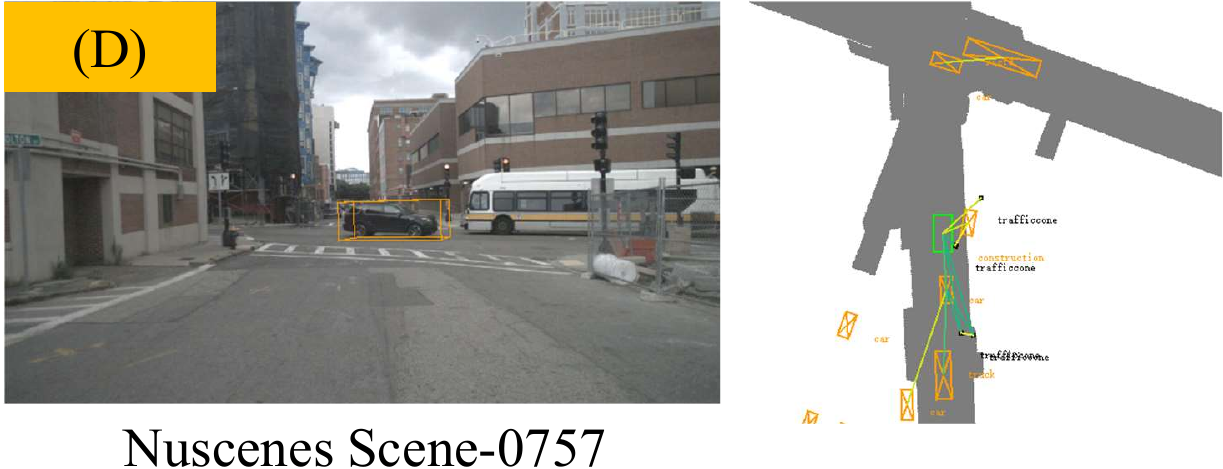}  
        \end{subfigure}
        \begin{subfigure}{.6\textwidth}
                \includegraphics[width=1\linewidth]{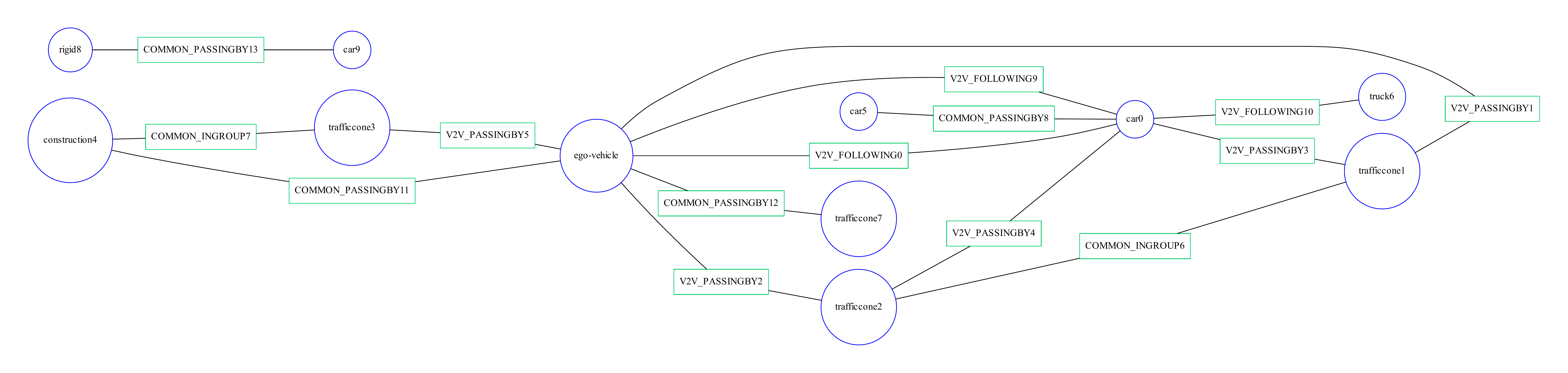}  
        \end{subfigure}
        \newline
        \begin{subfigure}{.39\textwidth}
                \includegraphics[width=1\linewidth]{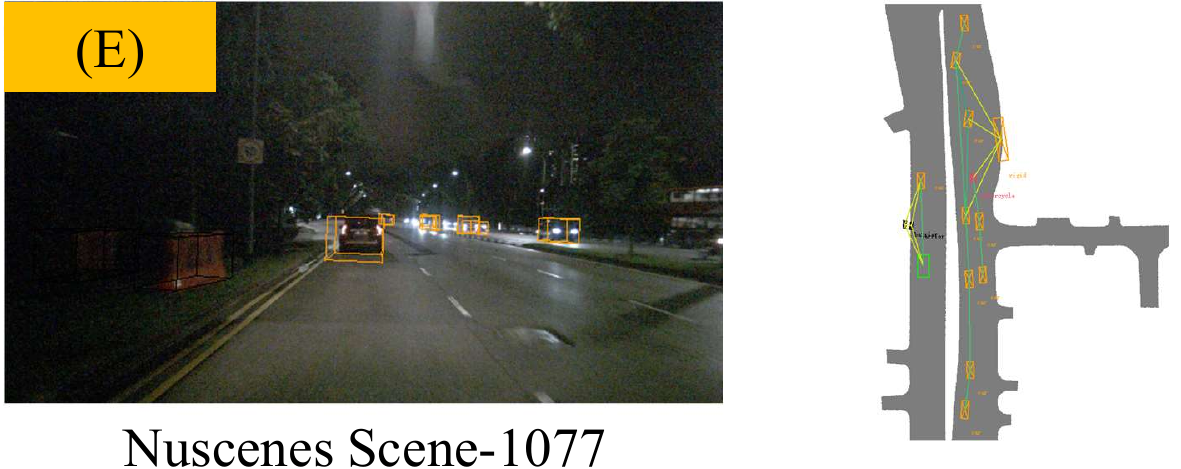}  
        \end{subfigure}
        \begin{subfigure}{.6\textwidth}
                \includegraphics[width=1\linewidth]{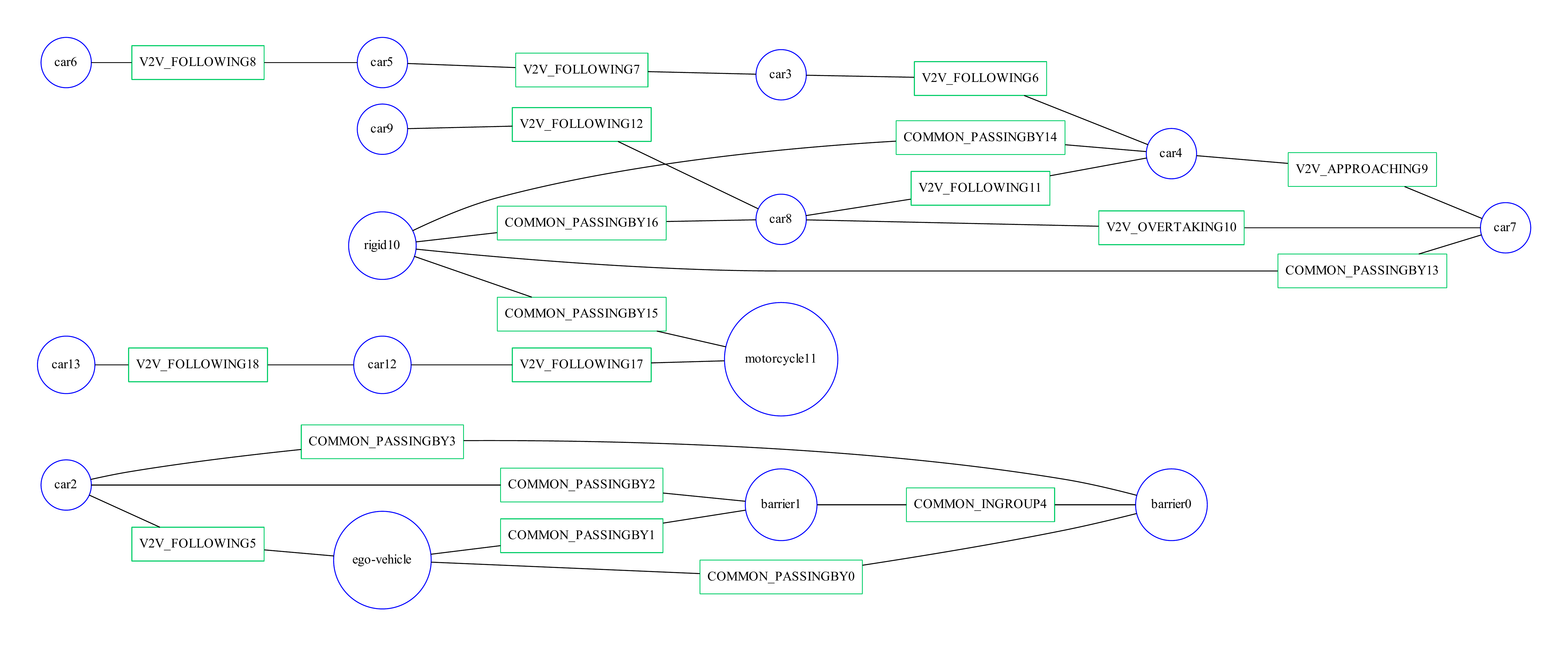}  
        \end{subfigure}
        \caption{Sample scenes in our dataset, front camera image (left), bird's-eye view (middle) and annotated scene graph (right).}
        \label{finalSample}
\vspace{-2em}
\end{figure*}

In this section, we describe how we setup and annotate the dataset. And then, we apply our graph generative model on road scene graph dataset to present graph refinement and graph generalization. Figure \ref{finalSample} propose some selected samples of road scene graph in nuScenes dataset.

\subsection{Data Setup}

\subsubsection{Data planning}

As Table~\ref{relatedDatasets} illustrates, there are several self-driving scene datasets with accurate geometry and rich semantic information. Among these datasets, nuScenes\cite{caesar2020nuscenes} dataset proposes significantly more annotations and object classes. Also, one major problem for the graph generative model is the non-unique node representation. And nuScenes dataset solved this problem by giving every entity an ID. So our research benefits from this measure, which is designed for better object tracking.

\subsubsection{Scene annotation}

\begin{figure}[htbp]
        \centering
        \includegraphics[width=0.8\linewidth]{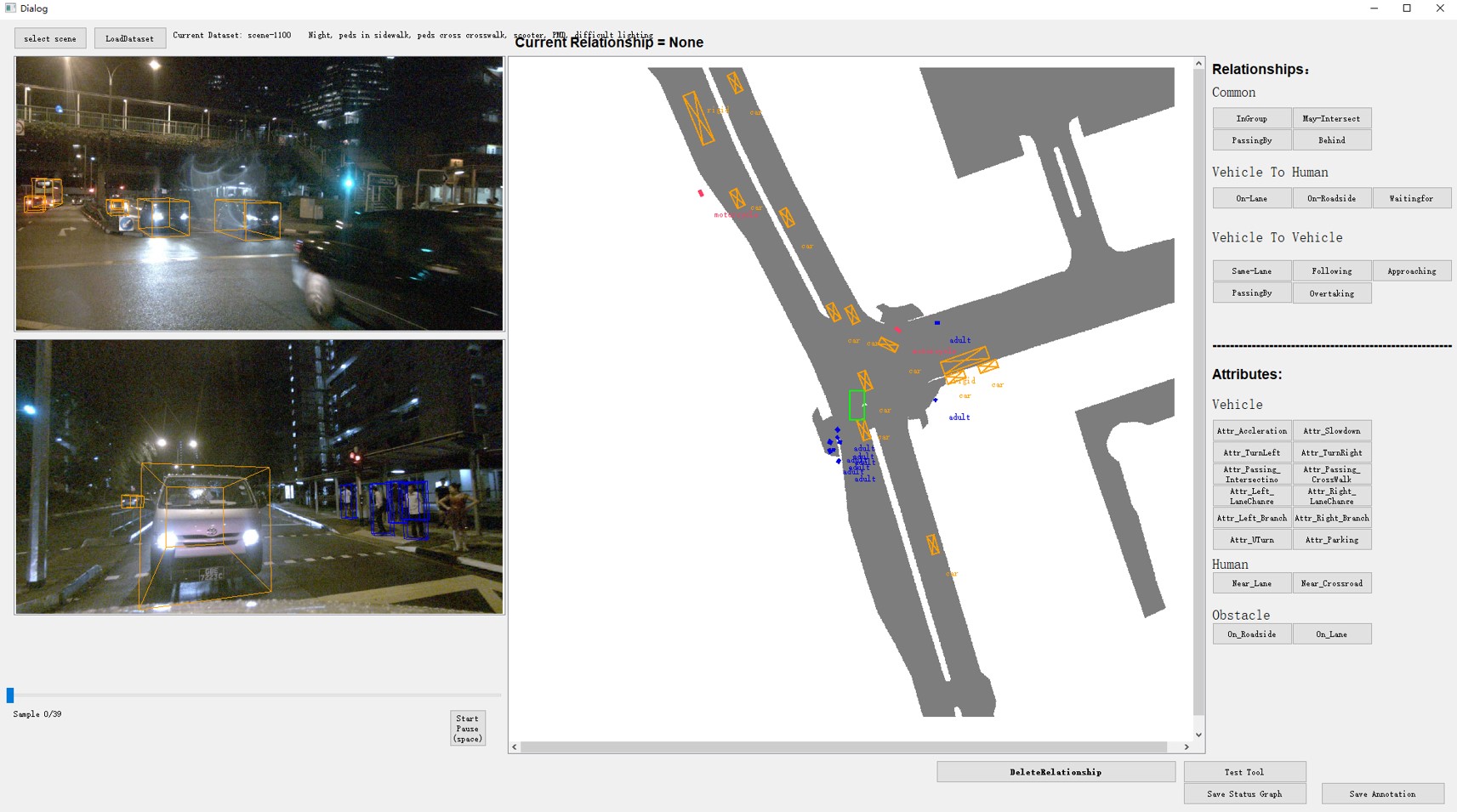}  
        \caption{Our data annotator for road scene graph dataset. }
        \label{DataAnnotatorImage}
\vspace{-1em}
\end{figure}

The first step of our research is the GUI data annotator, illustrated in Fig.~\ref{DataAnnotatorImage}. This program enables experienced drivers to annotate, create, and evaluate road scene graph from nuScenes and CARLA recorded data.

Another important question is: Which relationship is most important for road scene graph? For the relationships, we distribute questionnaires, along with the first 50 scenes from CARLA, to 5 drivers. And collect relationships that appear more 3 times, as Table~\ref{relationships} illustrates. "Group" relationship is a special case, which was added for better and faster annotation. And for the attributes, we borrow the concept of the common relationship from research \cite{ramanishka2018toward}. So our dataset will be compatible with their dataset.

\subsubsection{Annotation statistics}

\begin{figure}[htbp]
        \centering
        \includegraphics[width=0.8\linewidth]{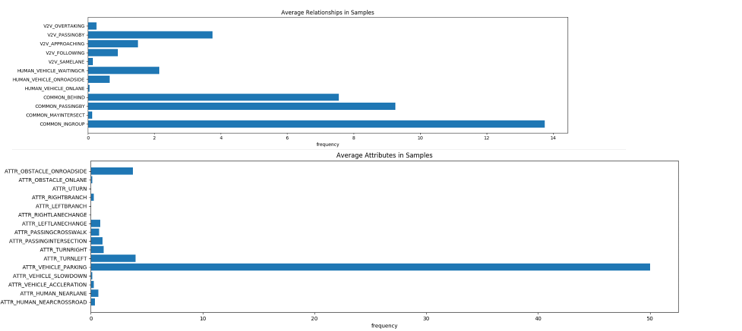}  
        \caption{Edge and attribute distribution in our dataset.}
        \label{distribution}
\vspace{-1em}
\end{figure}

Here we brought up some statistics about our road scene graph dataset. First of all,  edge and attribute distribution of scene graph illustrate in Fig.~\ref{distribution}. For relationship, the top 3 common relationships are : InGroup, PassingBy, Behind; So all of them are kinds of spatial-related relationships. And for attributes, the top 3 is Parking, TurnLeft, Obstacle On Roadside; As the vehicle parking at roadside are the most common objects in nuScenes Dataset.

\begin{figure}[htbp]
        \centering
        \includegraphics[width=0.8\linewidth]{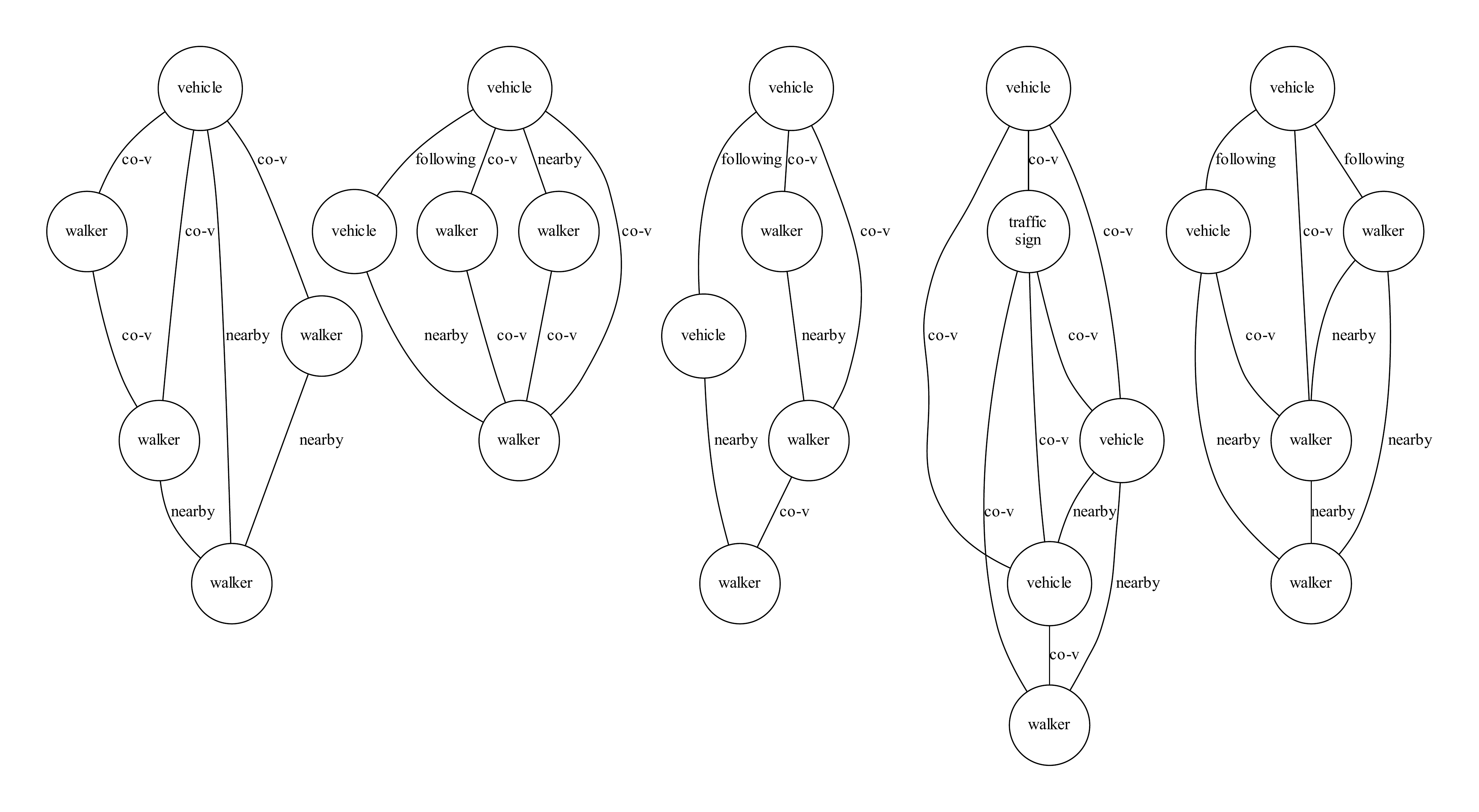}  
        \caption{The most common motifs in road scene graph dataset, corresponding to the most common object combinations among the dataset.} 
        \label{motifTop5}
\vspace{-1em}
\end{figure}

Then, to propose a more intuitive impression of this dataset, here we find out the top 5 $k=5$ most common graph motifs among the dataset, as Fig.~\ref{motifTop5} illustrates. This problem could be staged as an exceptional attributed subgraph mining problem. And we implemented the solution from research \cite{bendimerad2019mining}. These motifs are an excellent indicator of the high-frequency patterns in the dataset.
The top of it is a vehicle with a bunch of walkers nearby, And the fourth one is a vehicle and several walkers waiting for traffic lights. Which is quite a common pattern in the dataset.

\subsection{GCN Based road scene graph prediction network}

Here we propose the detail and performance of our graph refinement \& generation network. As currently, the size of our dataset is not sufficient to train a perfect graph generation network. We firstly implement a set of object pairwise relationship detector (for example, "following" detector) in CARLA. As Fig.~\ref{relationships} in Section~III illustrates, these detectors could only detect the star-marked relationship. And then, we use these datasets to construct a mini-road-scene-graph dataset in CARLA for pretraining. Then, we use nuScenes road scene graph to fine-tune the network. Which is very similar to our second task NGPGV (Next graph prediction using graph VGAE ). Here we propose the performance of our model. We adopted the scene-wise recall evaluation metrics R@k (R@5, R@15, R@25) from \cite{lu2016visual}.

\begin{table}[htbp]
        \caption{Performance on different models applied on specific tasks}
        \begin{tabular}{cc|cc}
        \multicolumn{2}{c}{}                                & No Pooling & Gated Pooling \\ \hline
        \multicolumn{1}{c|}{\multirow{2}{*}{RSGRN}}  & R@5  & 42.50      & 47.25         \\
        \multicolumn{1}{c|}{}                        & R@15 & 77.00      & 85.25         \\ \hline
        \multicolumn{1}{c|}{\multirow{2}{*}{NGPGV}}  & R@15 & 25.50      & 26.75         \\
        \multicolumn{1}{c|}{}                        & R@25 & 39.75      & 34.75         \\ \hline
        \multicolumn{1}{c|}{\multirow{2}{*}{NGPGVEB}} & R@15 & 13.50      & 11.00         \\
        \multicolumn{1}{c|}{}                        & R@25 & 35.75      & 25.00        
        \end{tabular}
        \centering
        \label{Prediction}
\vspace{-1em}
\end{table}

Comparing to the similar graph generation tasks in common objects and relationships dataset \cite{xu2017scene,xu2018pointfusion,lu2016visual}. Our graph generation model proposes a state-of-the-art performance in accuracy. And in the graph refinement task, our model also proposes a good accuracy. However, compared to the original road scene graph generation model (NGPGV), after we integrate image encoding inside the dataset. The performance obviously decreases by about 10 percent. It means currently our dataset cannot propose enough data for grounding objects into the image. Future research will be focusing on enlarging the dataset.

\section{CONCLUSION}
\label{s:conclusions}
In this paper, we propose road scene graph, which is a special scene graph which mainly focusing on intelligent vehicle's application. And it is initial research which applies topological graph-based data representation to intelligent vehicle. The main contribution of this paper is: 

\begin{itemize}
        \item  We firstly apply scene graph related methods on intelligent vehicle related areas. And proposed graph generative model to train this graph-based scene representation for road scenes.
        \item  To train our model, we propose a scene graph dataset for driving scenes, which is different from current common-relationship oriented scene graph dataset. 
        \item  Then, we propose a set of models for generating the road scene graphh. These generation model could significantly benefit the data annotation process. And proposing baseling model for further application like risk scene detection and scene captioning.
\end{itemize}

\subsection{Future works}

Future work will be focusing on two aspects: Current data amount is not sufficient for very complex semantic inference. Also, mainstream driving datasets include no risk/accident scene. So we would like to enlarge the dataset to about 1000 scenes, and adding an accident road scene graph from the Road Hazardous Stimuli dataset \cite{wolfe2020rapid} in 2021.

In the application aspect, future research will be focusing on scene captioning and risk scene classification. Which is a well-developed technique for a common-relationship scene graph.

\section*{Acknowledgments}
This work was supported by the Science Fund for Creative Research Groups of the National Natural Science Foundation of China under grant number 51521003. The research work carried out at the State Key Laboratory of Robotic and Intelligent System, Harbin Institute of Technology. We are deeply grateful for the kind cooperation of both faculty and students.

\bibliography{ref}

\end{document}